\documentclass[
]{ceurart}
\usepackage{makecell}

\usepackage{graphicx}
\graphicspath{ {./figures/} }
\begin{document}

\copyrightyear{2022}
\copyrightclause{Copyright for this paper by its authors.
  Use permitted under Creative Commons License Attribution 4.0
  International (CC BY 4.0).}

\conference{The International Conference on Agglutinative Language Technologies as a challenge of Natural Language Processing (ALTNLP), June 6 -- 8, Koper, Slovenia}

\title{Uzbek Sentiment Analysis based on local Restaurant Reviews}

\author[1]{Sanatbek Matlatipov}[%
orcid=0000-0002-6895-3436,
email=s.matlatipov@nuu.uz,
url=https://sanatbek.uz/,
]
\author[1]{Hulkar Rahimboeva}[%
orcid=0000-0002-3259-7708,
email=h.rahimboyeva@nuu.uz,
]
\author[1]{Jaloliddin Rajabov}[%
orcid=0000-0002-0369-6707,
email=j.rajabov@nuu.uz,
]
\author[2]{Elmurod Kuriyozov}[%
orcid=0000-0003-1702-1222,
email=e.kuriyozov@udc.es
]

\address[1]{National University of Uzbekistan named after Mirzo Ulugbek,
  4 Universitet St, Tashkent, 100174, Uzbekistan}
  
\address[2]{ Universidade da Coru\~na, CITIC, Grupo LYS, Depto. de Computaci\'on y Tecnologías de la Información, Facultade de Inform\'atica, Campus de Elvi\~na, A Coru\~na 15071, Spain
}
\begin{abstract}
Extracting useful information for sentiment analysis and classification problems from a big amount of user-generated feedback, such as restaurant reviews, is a crucial task of natural language processing, which is not only for customer satisfaction where it can give personalized services, but can also influence the further development of a company. 
In this paper, we present a work done on collecting restaurant reviews data as a sentiment analysis dataset for the Uzbek language, a member of the Turkic family which is heavily affected by the low-resource constraint, and provide some further analysis of the novel dataset by evaluation using different techniques, from logistic regression based models, to support vector machines, and even deep learning models, such as recurrent neural networks, as well as convolutional neural networks.
The paper includes detailed information on how the data was collected, how it was pre-processed for better quality optimization, as well as experimental setups for the evaluation process.
The overall evaluation results indicate that by performing pre-processing steps, such as stemming for agglutinative languages, the system yields better results, eventually achieving 91\% accuracy result in the best performing model.
\end{abstract}


\maketitle

\section{Introduction}
\label{sec:introduction}
The power of Natural Language Processing (NLP) techniques relies on amounts of labelled data in many applications. Sentiment analysis is the process of analyzing and labelling the opinion which is posted by consumers.  Consumers usually post their feedback about places/foods to famous applications such as Google Maps \footnote{Google Maps: \url{https://www.google.com/maps}}, Yelp\footnote{Yelp: \url{https://www.yelp.com}}, etc). They often encourage consumers to actively participate in reviews, and massive user-generated restaurant reviews allow consumers to fully express their needs while helping merchants provide real-time and personalized service \cite{anayasanchez19}. Moreover, the restaurant reviews express the composition of clients' emotional necessities and are an important source of information about consumers' choices~\cite{Marine-Roig2015-iz}. Currently, opinion mining has achieved very high accuracy performances, especially after applying deep learning methods, for high resource languages \cite{barnes2017assessing}. However, applying deep learning and machine learning techniques for different types of domains \cite{Zhang18} and gathering corpora with high quantity play an important\cite{https://doi.org/10.48550/arxiv.2203.08111} role in the development of low-resource languages. For example, the language we focus on is the Uzbek language which is being used by around 34 million native speakers in Uzbekistan and elsewhere in Central Asia and China\footnote{\url{https://en.wikipedia.org/wiki/Uzbek_language}}. Uzbek is a null-subject and highly agglutinative language where one word can be a meaningful sentence\cite{matlatipov2009representation, 10.1007/978-3-030-63119-2_59}. To our knowledge, there is no previous work for sentiment classification problems based on restaurant domain feedback. So, the following contributions are considered for this paper.
\begin{itemize}
\item \verb|Restaurant domain annotated corpora| is created for sentiment analysis which is collected from Google Maps based on Uzbek cuisine's locations where local national food reviews are the primary target. The corpora contain 4500 positive and 3710 negative reviews after manually removing major errors and cleaning. The annotation process is based on the feedback's 5 stars method provided by Google Maps where from 1 to 3 we consider the dataset as negative and from 4 to 5 as positive. We found some reviews are based on other languages such as English, Kyrgyz and Russian. We didn't want to ignore them, so we decided to translate them into Uzbek using the official Google Translate API. 
\item \verb|Pre-processing the corpora| is applied in two steps. The first steps are removing URLs, punctuation, and lower-casing. The second step is ignoring stopwords\cite{Madatov2022-el} from the dataset where it is based on accuracy evaluation after generating the list of stop words using the TF-IDF algorithm; Then, we applied the stemming algorithm \cite{10.1007/978-3-030-63119-2_59, 8747021} which is based on Uzbek words’ endings’ electronic dictionary that uses combinatorial approach inferring apply for part of speech of the Uzbek language: nouns, adjectives, numerals, verbs, participles, moods, voices. Advantages of using the algorithm are lexicon-free and its complexity that allows one operation (referring to the dictionary of endings of the language) to perform: segmentation of the word into suffixes; performs morphological analysis of the word.
\item \verb|Machine learning and deep learning| algorithms have been applied. Furthermore, deep learning(Recurrent neural network)  algorithm fed with fastText\footnote{\url{https://fasttext.cc/docs/en/crawl-vectors.html}} pre-trained word embedding is applied to improve the accuracy; 
\end{itemize}
All resources including the corpora, source code used for crawling techniques and classification algorithms are uploaded to the public repository \footnote{\url{https://github.com/SanatbekMatlatipov/restauranat-sentiment/tree/main}}.
The paper is structured as follows:
Introduction(this section), Section \ref{sec:related} describes related work that has been done so far. It is followed by a description of the methodology in Section \ref{sec:methodology} and continues with Section \ref{sec:experiemnts}, which focuses on experiments and results. The final part (Section \ref{sec:conclusion}) concludes the paper and highlights the future work.

\section{Related Work}
\label{sec:related}
In recent years, several works were done in the NLP field for Uzbek, including sentiment analysis datasets~\cite{rabbimov2021opinion, kuriyozov2019building}, created by collecting and analyzing Google Play app reviews, with two types of data: A medium-size manually annotated dataset and a larger-size dataset automatically translated from English. \cite{kuriyozov-etal-2020-cross} obtained bilingual dictionaries for six Turkic languages and applied them to cross-lingually align word embeddings, backed by a bilingual dictionary induction evaluation task. They showed that obtained aligned word embeddings from a low-resource language can benefit from resource-rich closely-related languages. Another similar paper \cite{10.1007/978-3-030-60276-5_42} investigated the effect of emoji-based features in opinion classification of Uzbek texts. A semantic evaluation dataset was presented with semantic similarity and relatedness scores in word pairs as well as its analysis for Uzbek in a recent work \cite{salaev2022simreluz}. There is a very recent growing trend in NLP that makes use of AI-based techniques, which can be seen in the work on Uzbek with neural transformers-architecture based language model trained off raw Uzbek corpus \cite{mansurov2021uzbert}.

In a global outlook to the field of sentiment analysis, there is a work~\cite{9083703} that used various methods of sentiment analysis techniques, such as machine learning and deep learning, in their work with an idea to take into account the differences in opinions and thoughts that exist on popular social platforms such as Twitter, Reddit, Tumblr and Facebook.

\section{Methodology}
\label{sec:methodology}
\begin{figure}
  \centering
  \includegraphics[width=0.7\linewidth]{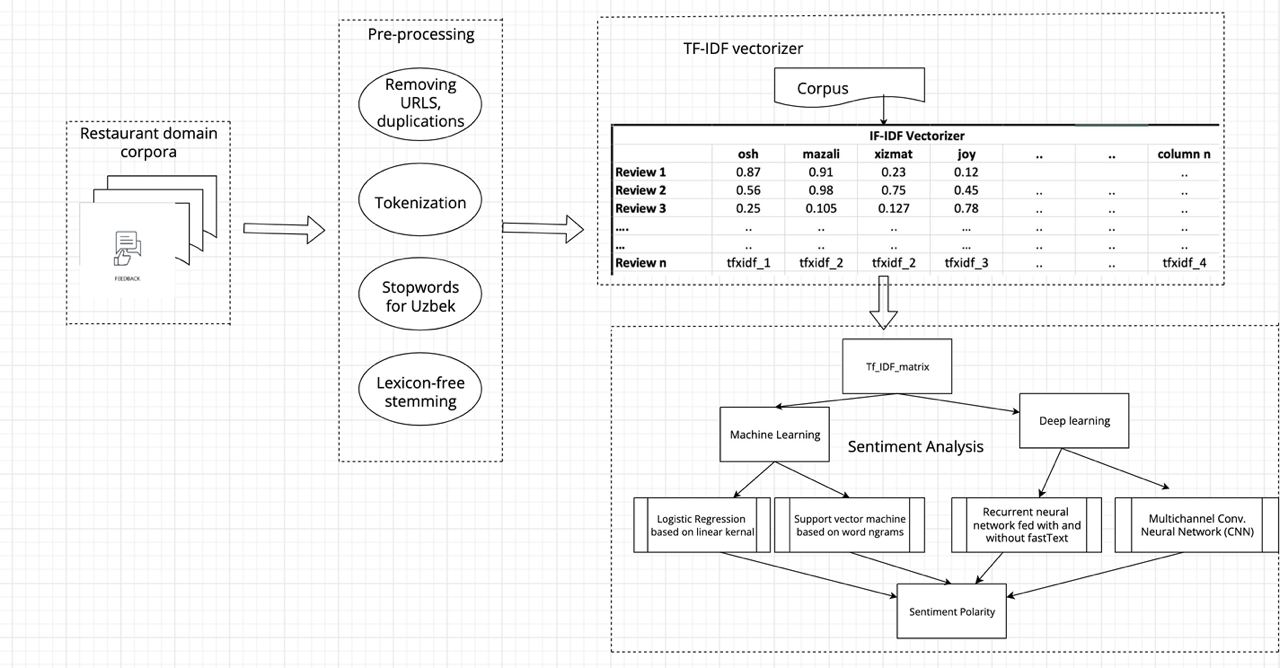}
  \caption{Research framework.}
  \label{fig:research_framework}
\end{figure}

In this paper, we proposed a machine learning and deep learning-based sentiment analysis framework for the restaurant domain dataset (Figure \ref{fig:research_framework}). The framework includes data collection using web-crawler, pre-processing(cleaning, stopwords, lexicon-free stemming), constructing TF-IDF weight matrix, performing ML and DL for sentiment analysis;  

\subsection{Data collection}
\begin{figure}
  \centering
  \includegraphics[width=0.7\linewidth]{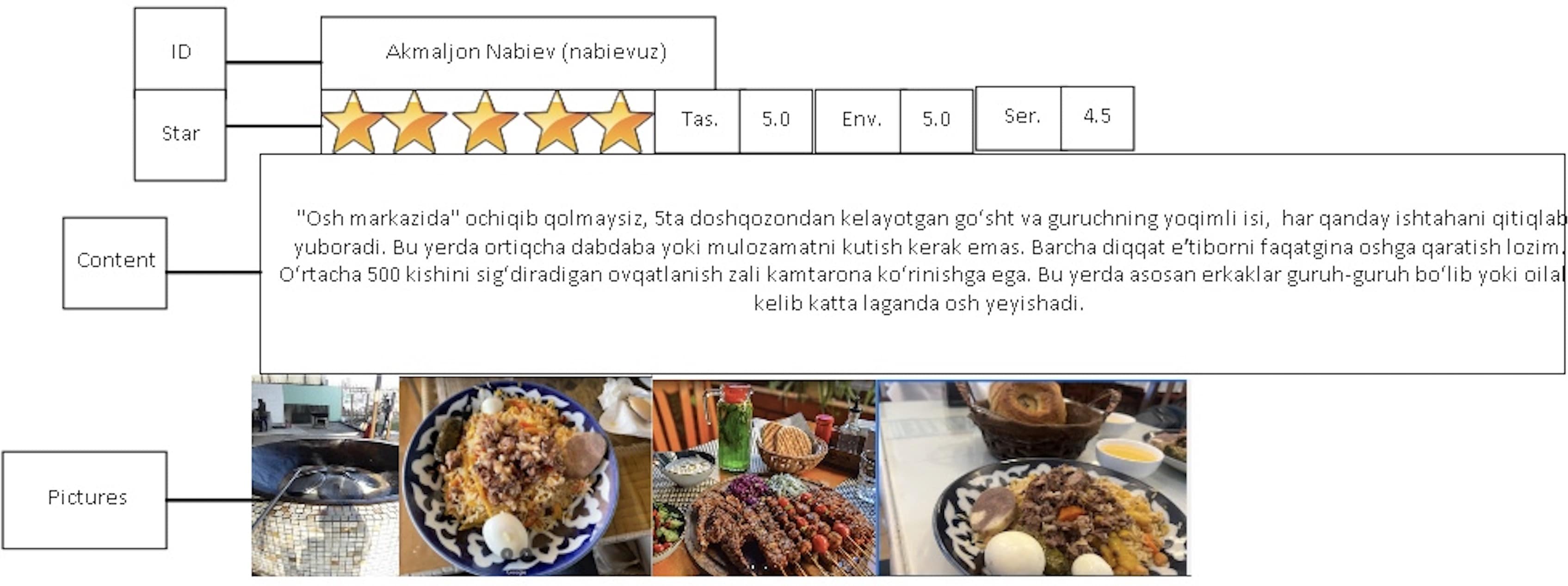}
  \caption{Feedback sample}
  \label{fig:review_sample}
\end{figure}

We start by looking at a high number of the dataset available for crawling in the Uzbek language. However, the usual approaches such as Twitter or movie reviews are not the case for Uzbek. Therefore, we decided to collect restaurant reviews as local people mostly loved giving feedback which is restaurants. we think it makes sense as Uzbek cuisines are one of the most popular throughout the Commonwealth of Independent States (CIS, CA countries). In most CA cities, for instance, it’s easy to find busy restaurants specializing in Uzbek cuisine\footnote{BBc Travel: \url{https://www.bbc.com/travel/article/20191117-is-uzbek-cuisine-actually-to-die-for}}. We crawled all local restaurants in Tashkent from Google Maps. Firstly, we selected a list of more than 140 URLs which has at least 3 reviews and we retrieved all info shown in Figure~\ref{fig:review_sample}.
While crawling, we considered Google’s anti-spam and anti-DDOS policies as there
are certain limitations on harvesting data. The source code is available on the repository.

\subsection{Data pre-processing}
The collection of texts with star ratings in the crawled dataset was noisy and required manual correction. 
The comments containing only emojis, names or any other irrelevant content, such as username mentions, URLs or specific app names were removed. Those written in languages different from Uzbek (mostly in Russian and some in English) were translated using the official Google translate API. Although people in Uzbekistan use  the official Latin alphabet, the use of the old Cyrillic alphabet is equally popular, especially among adults. The comments that were written in Cyrillic were converted to Latin using the Uzbek machine transliteration tool \cite{salaev2022machine}. Then, we applied stop words to remove low-level information words from our comments to focus on important information. The technique is based on \cite{Madatov2022-el} paper where it is a proposed algorithm of automatic detection of single word stop words collection using TFIDF(Term frequency - inverse document frequency). After that, each word is processed to lexicon-free stemming tool \cite{10.1007/978-3-030-63119-2_59} algorithm for decreasing the word capacity because of prefixes and suffixes. The basic idea is using the combinatorial approach of eligible endings candidates.  Following table \ref{tab:processed_data} shows processed data which is ready for TFIDF-vectorizer. 
\begin{table}[!ht]
    \caption{The example of a chosen review before and after processing it. }
    \label{tab:processed_data}
    \centering
    \begin{tabular}{p{0.6\columnwidth}p{0.4\columnwidth}}
        \multicolumn{1}{c}{\textbf{Review}} &
        \multicolumn{1}{c}{\textbf{After processing}} \\ \hline
        Birinchi Milliy taomlardan biri - keng assortimentli taomlar! Gastro-turistlar uchun juda jozibali joy - bu yerda barcha turdagi milliy taomlar mavjud.Yagona salbiy tomoni shundaki, bunday yirik muassasa uchun to'xtash joyi kichik. Narxlar nisbatan arzon! Turistlar uchun juda arzon! &
        Bir/ milliy/ taom/ keng/ assortiment/ taom/ gastro/ turist/ juda/ joziba/ joy/ tur/ milliy/ taom/ mavjud/ salbiy/ tomon/ yirik/ muassa/ to’xta/ joy/ kichik/ narx/ arzon/ turist/ arzon
    \end{tabular}
\end{table}

\begin{figure}[!ht]
  \centering
  \includegraphics[width=0.6\columnwidth]{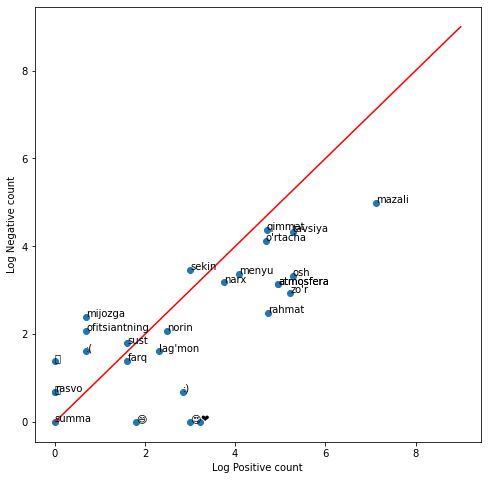}
  \caption{The visualisation of some selected examples of Uzbek words taken from positive and negative reviews with their log counts.}
  \label{fig:log_count}
\end{figure}

We selected a set of words to visualize the word count. Figure \ref{fig:log_count} shows that people tend to give more positive feedback than negative on the domain of restaurants.

\section{Evaluation}
\label{sec:experiemnts}
The collected novel dataset has been split into training and testing subsets for evaluation with 8 x 2 ratio respectively.
After the data cleaning process, we have the original dataset as follows, where $\vec{x_{i}}$ represents feature vectors and $\vec{y_{i}}$ represents annotated labels:

\begin{equation}
  (\vec{x_{i}}, y_{i}), \hspace{2cm} i = 1,2,3,...,N
\end{equation}
\begin{equation}
  \vec{x_{i}} = (x_{i1}, x_{i2}, ... , x_{im}) \hspace{1cm} i = 1,2,3,...,N 
\end{equation}
$N$ and $m$ is equal to the number of reviews and length of the feature vector, respectively.

Then we calculate TFIDF scores for each feature vector $\vec{x_{i}}$ which vectorises words by taking into account the frequency of a word in a given review and the frequency between reviews. The final result of all $\vec{z_{i}}$s is defined as a sparse matrix.
\begin{equation}
  \vec{z_{i}} = TF(x_{i}) x IDF(x_{i}) \hspace{1cm} i = 1,2,3,...,N 
\end{equation}

\subsection{Machine learning algorithms}
The \verb|\Logistic regression| model is
\begin{equation}
  h(\vec{z}) = 1 / (1 + \exp({-z})) 
\end{equation}
\[
    P(y|\vec{z})= 
\begin{cases}
    h(\vec{z}),& \text{if } y = +1(positive)\\
    1-h(\vec{z}), & \text{if } y = -1(negative)
\end{cases}
\]
Logistic regression\cite{CHRISTODOULOU201912} model is a classification algorithm, known for its exponential and log-linear functions. It works with discrete values and maps the function of any real value into 0 and 1. For sentiment analysis, the hypothesis shows, reviews are either positive or negative by using the (4).
The Support Vector Machine(SVM) model has the following response function:
\begin{equation}
  h(\vec{z}) = sign(\vec{z}) 
\end{equation}
SVM algorithm is known for its fast and dependable classification which resolves two-group classification problems. The classification is conducted for finding a hyperplane between two classes’ positive and negative reviews in the model:
After all, we implemented LR and SVM models utilizing the Scikit-Learn \cite{scikit-learn} machine learning library in Python with default configuration parameters. 
For the LR models, we implemented a variant based on word n-grams (unigrams and bigrams), and 
one
with character n-grams (with $n$ ranging from 1 to 4).
We also tested a model combining 
said
word and character n-gram features.

\subsection{Deep Learning algorithms}
\begin{figure}[!ht]
  \centering
  \includegraphics[width=0.6\columnwidth]{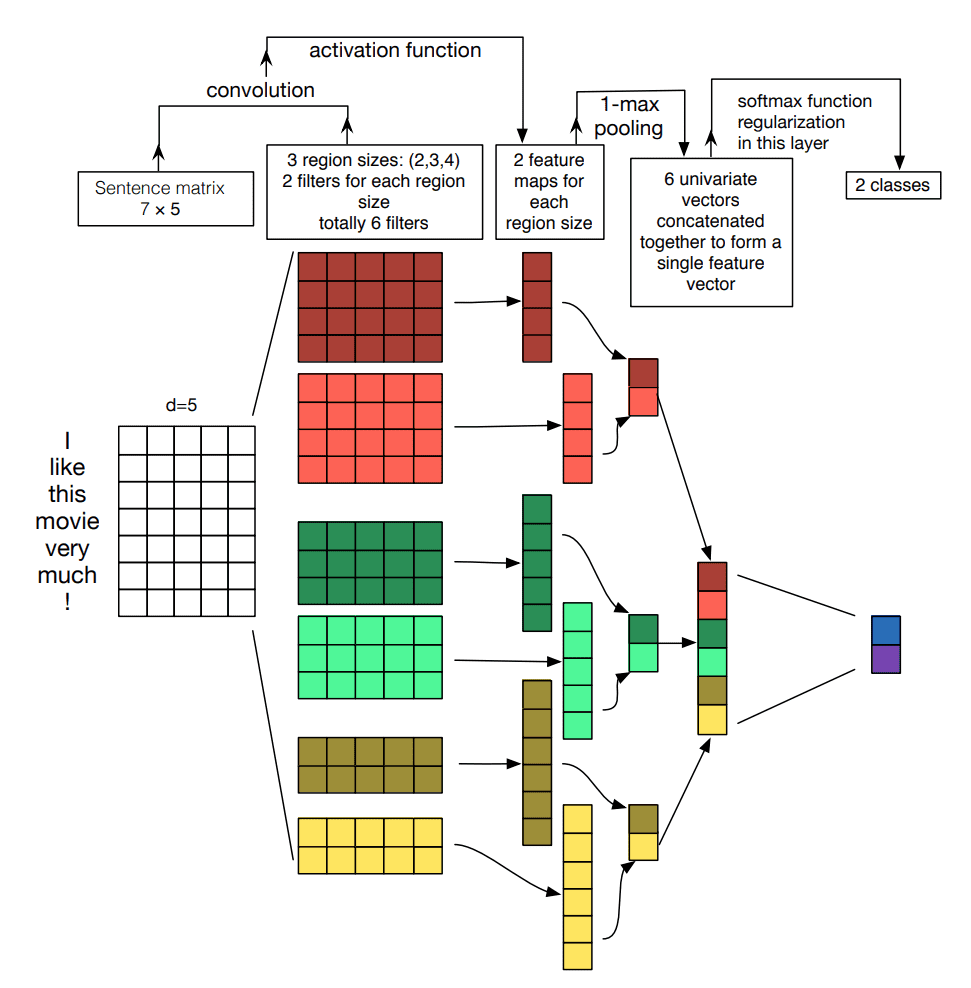}
  \caption{The illustration of steps taken in deep learning based sentiment classification using CNN, inspired by \cite{zhang2015sensitivity}.}
  \label{figure:visualisation-cnn}
\end{figure}

Keras ~\cite{chollet2015keras} is used
on top of TensorFlow~\cite{tensorflow2015-whitepaper-abbr}.The FastText pre-trained word embeddings of size 300~\cite{grave2018learning} for the Uzbek language are applied.  For the CNN model, we used a multi-channel CNN with 256 filters and three parallel channels with kernel sizes of 2,3 and 5, and drop out of 0.3. The output of the hidden layer is the concatenation of the max-pooling of the three channels.  For RNN, we use a bidirectional network of 100 GRUs. The output of the hidden layer is the concatenation of the average and max-pooling of the hidden states. For the combination of deep learning models,  we stacked the CNN on top of the GRU. In the three cases, the final output is obtained through a sigmoid activation function~\cite{marreiros2008population} applied to the previous layer. In all cases, Adam optimization algorithm \cite{kingma2014adam}, an extension of stochastic gradient descent, was chosen for training, with standard parameters: learning rate $\alpha=0.0001$ and exponential decay rates $\beta_1 =0.9$ and $\beta_2 =0.999$. Binary cross-entropy was used as a loss function.
The same steps, but slightly different parameters were used in a work that presents guidance to use CNN for sentiment classification \cite{zhang2015sensitivity}. Inspired by their example that perfectly illustrates the steps of performing deep learning based sentiment classification using CNN, the visualisation of our steps can be seen in Figure \ref{figure:visualisation-cnn}.

\subsection{Evaluation metrics}
Confusion\cite{SOKOLOVA2009427} matrices are used in the task to determine the gap between predicted and true values which is shown in Table \ref{tab:title}. Precision, Recall and F1-score are used as evaluation metrics for model performance.

\begin{table}[!ht]
\caption{Confusion matrix}
\label{tab:title} 
\centering
\begin{tabular}{lll}
\multicolumn{1}{c}{\textbf{Classes}} & \multicolumn{1}{c}{\textbf{Positive}} & \multicolumn{1}{c}{\textbf{Negative}} \\ \hline
\textbf{Positive}                    & True Positive(TP)                     & False Negative(FN)                    \\
\textbf{Negative}                    & False Positive(TP)                    & True Negative(FN)                    
\end{tabular}
\end{table}

The calculation of Precision and Recall is shown below:
\begin{equation}
  Precision = \frac{TP}{TP + FP} \hspace{1cm} Recall = \frac{TP}{TP + FN}
\end{equation}
The F1-Score is used, which takes into account both accuracy and recall, and the F1-Score is
calculated as follows:
\begin{equation}
  F_{1} = \frac{2 * Precision * Recall}{Precision + Recall}
\end{equation}

\section{Results and Discussion}
This section presents a detailed description of the results obtained by the evaluation process using both machine learning and deep learning techniques applied to the collected novel sentiment analysis dataset of restaurant reviews.

\subsection{Experiment Results}
The overall experiment results of the above-mentioned evaluation were performed, and the results can be seen in Table \ref{table:results}.

\begin{table}[!ht]
\caption{Experiment results of sentiment analysis for all evaluation techniques, including models, their distinctive parameters, as well as evaluation metrics, such as Precision (Prec.), Recall (Rec.), F1-score (F1), as well as Accuracy (Acc.).}
\label{table:results}
\centering
\begin{tabular}{llrrrr}
\textbf{Model name + Parameters} &
  \textbf{Sentiment} &
  \multicolumn{1}{l}{\textbf{Prec.}} &
  \multicolumn{1}{l}{\textbf{Rec.}} &
  \multicolumn{1}{l}{\textbf{F1}} &
  \multicolumn{1}{l}{\textbf{Acc.}} \\ \hline
\multirow{2}{*}{Logistic Regression based on word n-grams}  & Positive & 88\% & 98\% & 93\% & \multirow{2}{*}{89\%} \\
                                                           & Negative & 88\% & 67\% & 74\% &                       \\
\multirow{2}{*}{Logistic Regression based on char. n-grams} & Positive & 87\% & 51\% & 92\% & \multirow{2}{*}{87\%} \\
                                                           & Negative & 83\% & 97\% & 64\% &                       \\
\multirow{2}{*}{Logistic Regression (word + char. n-grams)} & Positive & 95\% & 95\% & 92\% & \multirow{2}{*}{91\%} \\
                                                           & Negative & 90\% & 89\% & 90\% &                       \\
\multirow{2}{*}{SVM based on linear kernel}                & Positive & 88\% & 97\% & 92\% & \multirow{2}{*}{88\%} \\
                                                           & Negative & 84\% & 71\% & 80\% &                       \\
\multirow{2}{*}{RNN without word embeddings}               & Positive & 90\% & 95\% & 92\% & \multirow{2}{*}{88\%} \\
                                                           & Negative & 78\% & 64\% & 70\% &                       \\
\multirow{2}{*}{RNN with word embeddings}                  & Positive & 90\% & 95\% & 93\% & \multirow{2}{*}{88\%} \\
                                                           & Negative & 80\% & 65\% & 72\% &                       \\
\multirow{2}{*}{CNN (multichannel)}                        & Positive & 90\% & 96\% & 93\% & \multirow{2}{*}{89\%} \\
                                                           & Negative & 83\% & 64\% & 72\% &                      \\ \hline
\end{tabular}
\end{table}

The Logistic Regression(LR) based on word n-grams obtained a binary classification accuracy of 90\% on the dataset, while the one based on character n-grams, with its better handling of misspelt words, improved it to 91\%(which is the winner of this comparison). Support Vector machines based on Linear kernel mode have shown 88\% accuracy overall. Recurrent Neural network models without and with fastText embedding show the same accuracy (88\%). Convectional Neural Network showed slightly less performance(89.23\%)  than LR. However, this is the reason for lacking data for neural-network models, as it requires big data for better performance.

\subsection{Discussion and limitations}
Nowadays, unstructured data are becoming more and more in the restaurant domain which requires performing high accuracy sentiment analysis. Especially, this is the case for low-resource languages. Based on the review data of Google Maps(Tashkent location) which is obtained by web-crawling, the paper has shown several ML\& DL methods. It was observed that the LR algorithm outperforms the others which makes sense as our dataset is relatively small. The research also mentioned some theoretical and practical implications. We believe, in terms of gaining massive user reviews on the domain can provide consumers make their decision in the best manner such as lower cost and faster speed.
However, we also wanted to point out some limitations in this research paper. The dataset we gathered has an unbalanced number of positive and negative reviews, which can cause deviations in the result. Moreover, we used the review rating in the annotation process which sometimes, in reality, consumers may give a high rating score, but polarity context may be related to negative, and vice versa.

\section{Conclusion}
\label{sec:conclusion}
In this paper, we have shown a novel dataset in the restaurant domain for the Uzbek language, with 8210 reviews, annotated with positive or negative labels, which is crawled from Google Maps using URLs of all locations in the capital city Tashkent, and was labelled as their corresponding star score. Then, we applied full pre-processing steps to the dataset which contributed to increasing the accuracy of our baseline models. Further analysis of the collected dataset was shown with evaluations using both machine learning and deep learning techniques. The best accuracy result (91\%) on the dataset was obtained using a logistic regression model with word and character n-grams.

In the foreseen future, we are planning to extend the work by collecting more data, which can effectively analyze the restaurant reviews in a practical level. Also, the work is underway to remove the evaluation bias of the training experiments by using cross-validation methods in data splitting.

\begin{acknowledgments}
This work partially has received funding from ERDF/MICINN-AEI (SCANNER-UDC, PID2020-113230RB-C21), and from Centro de Investigación de Galicia ''CITIC'', funded by Xunta de Galicia and the European Union (ERDF - Galicia 2014-2020 Program), by grant ED431G 2019/01. Elmurod Kuriyozov was funded for his PhD by El-Yurt-Umidi Foundation under the Cabinet of Ministers of the Republic of Uzbekistan.
\end{acknowledgments}

\bibliography{main}

\end{document}